\documentclass[runningheads]{llncs}
\usepackage{graphicx}
\usepackage{amsmath,amssymb} 
\usepackage{color}
\usepackage[width=122mm,left=12mm,paperwidth=146mm,height=193mm,top=12mm,paperheight=217mm]{geometry}
\begin{document}
\pagestyle{headings}
\mainmatter

\title{TextSnake: A Flexible Representation for Detecting Text of Arbitrary Shapes} 

\titlerunning{TextSnake}

\authorrunning{Shangbang Long et al.}

\author{Shangbang Long\textsuperscript{1,2}, Jiaqiang Ruan \textsuperscript{1,2}, Wenjie Zhang\textsuperscript{1,2}, Xin He\textsuperscript{2}, Wenhao Wu\textsuperscript{2}, Cong Yao\textsuperscript{2}}

\institute{\textsuperscript{1}Peking University, \textsuperscript{2}Megvii (Face++) Technology Inc.\\
	\email{ \{longlongsb, jiaqiang.ruan, zhang\_wen\_jie\}@pku.edu.cn, \{hexin,wwh\}@megvii.com, yaocong2010@gmail.com}
}


\maketitle

\begin{abstract}
Driven by deep neural networks and large scale datasets, scene text detection methods have progressed substantially over the past years, continuously refreshing the performance records on various standard benchmarks. However, limited by the representations (axis-aligned rectangles, rotated rectangles or quadrangles) adopted to describe text, existing methods may fall short when dealing with much more free-form text instances, such as curved text, which are actually very common in real-world scenarios. To tackle this problem, we propose a more flexible representation for scene text, termed as \textit{TextSnake}, which is able to effectively represent text instances in horizontal, oriented and curved forms. In TextSnake, a text instance is described as a sequence of ordered, overlapping disks centered at symmetric axes, each of which is associated with potentially variable radius and orientation. Such geometry attributes are estimated via a Fully Convolutional Network (FCN) model. In experiments, the text detector based on TextSnake achieves state-of-the-art or comparable performance on Total-Text and SCUT-CTW1500, the two newly published benchmarks with special emphasis on curved text in natural images, as well as the widely-used datasets ICDAR 2015 and MSRA-TD500. Specifically, TextSnake outperforms the baseline on Total-Text by more than \textit{$40\%$} in F-measure.

\end{abstract}

\section{Introduction}

In recent years, the community has witnessed a surge of research interest and effort regarding the extraction of textual information from natural scenes, a.k.a. scene text detection and recognition. The driving factors stem from both application prospect and research value. On the one hand, scene text detection and recognition have been playing ever-increasingly important roles in a wide range of practical systems, such as scene understanding, product search, and autonomous driving. On the other hand, the unique traits of scene text, for instance, significant variations in color, scale, orientation, aspect ratio and pattern, make it obviously different from general objects. Therefore, particular challenges are posed and special investigations are required. 

\begin{figure*}
\begin{centering}
\vspace{-0mm}
\includegraphics[width=0.9\columnwidth]{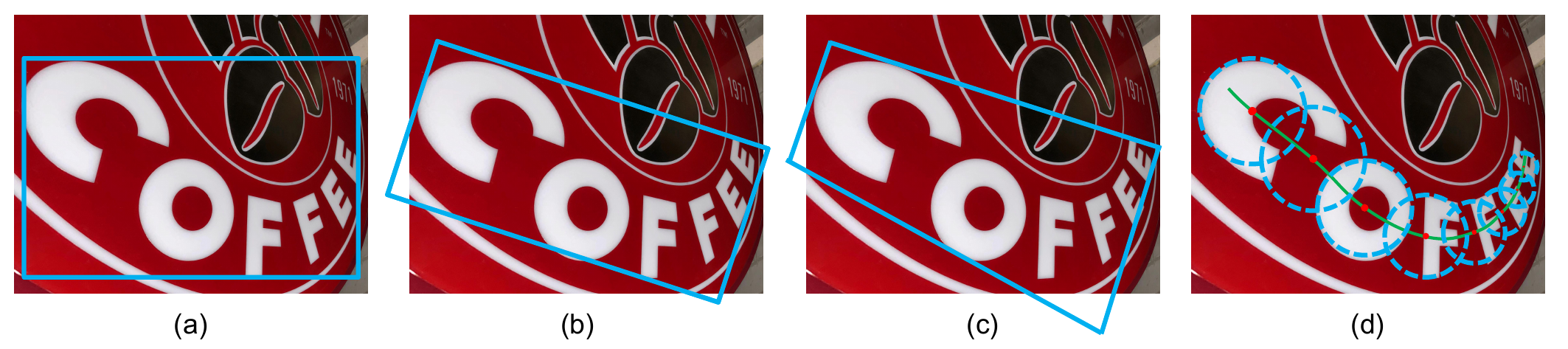}
\par\end{centering}
\vspace{-5mm}
\caption{Comparison of different representations for text instances. (a) Axis-aligned rectangle. (b) Rotated rectangle. (c) Quadrangle. (d) TextSnake. Obviously, the proposed TextSnake representation is able to effectively and precisely describe the geometric properties, such as location, scale, and bending of curved text with perspective distortion, while the other representations (axis-aligned rectangle, rotated rectangle or quadrangle) struggle with giving accurate predictions in such cases.}   \label{fig:representations}
\vspace{-8mm}
\end{figure*}

Text detection, as a prerequisite step in the pipeline of textual information extraction, has recently advanced substantially with the development of deep neural networks and large image datasets. Numerous innovative works~\cite{yao2016scene,Shi_2017_CVPR,Zhou_2017_CVPR,liao2017textboxes,huang2015densebox,wu2017self,he2017multi,Hu_2017_ICCV,tian2017wetext,Lyu2018,ZhangAAAI2018} are proposed, achieving excellent performances on standard benchmarks.


However, most existing methods for text detection shared a strong assumption that text instances are roughly in a linear shape and therefore adopted relatively simple representations (axis-aligned rectangles, rotated rectangles or quadrangles) to describe them. Despite their progress on standard benchmarks, these methods may fall short when handling text instances of irregular shapes, for example, curved text. As depicted in Fig.~\ref{fig:representations}, for curved text with perspective distortion, conventional representations struggle with giving precise estimations of the geometric properties.

In fact, instances of curved text are quite common in real life~\cite{kheng2017total,Yuliang2017Detecting}. In this paper, we propose a more flexible representation that can fit well text of arbitrary shapes, i.e., those in horizontal, multi-oriented and curved forms. This representation describes text with a series of ordered, overlapping disks, each of which is located at the center axis of text region and associated with potentially variable radius and orientation. Due to its excellent capability in adapting for the complex multiplicity of text structures, just like a snake changing its shape to adapt for the external environment, the proposed representation is named as TextSnake. The geometry attributes of text instances, i.e., central axis points, radii and orientations, are estimated with a single Fully Convolutional Network (FCN) model. Besides ICDAR 2015 and MSRA-TD500, the effectiveness of TextSnake is validated on Total-Text and SCUT-CTW1500, which are two newly-released benchmarks mainly focused on curved text. The proposed algorithm achieves state-of-the-art performance on the two curved text datasets, while at the same time outperforming previous methods on horizontal and multi-oriented text, \textit{even in the single-scale testing mode}. Specifically, TextSnake achieves significant improvement over the baseline on Total-Text by \textit{$40.0\%$} in F-measure.

In summary, the major contributions of this paper are three-fold: (1) We propose a flexible and general representation for scene text of arbitrary shapes; (2) Based on this representation, an effective method for scene text detection is proposed
; (3) The proposed text detection algorithm achieves state-of-the-art performance on several benchmarks, including text instances of different forms (horizontal, oriented and curved).

\section{Related Work}

In the past few years, the most prominent trend in the area of scene text detection is the transfer from conventional methods~\cite{epshtein2010detecting,neumann2010method} to deep learning based methods~\cite{jaderberg2014deep,jaderberg2016reading,liao2017textboxes,Zhou_2017_CVPR,Shi_2017_CVPR}. In this section, we look back on relevant previous works. For comprehensive surveys, please refer to~\cite{ye2015text,zhu2016scene}.  Before the era of deep learning, SWT~\cite{epshtein2010detecting} and MSER~\cite{neumann2010method} are two representative algorithms that have influenced a variety of subsequent methods~\cite{yin2014robust,huang2014robust}. Modern methods are mostly based on deep neural networks, which can be coarsely classified into two categories: regression based and segmentation based.

Regression based text detection methods~\cite{liao2017textboxes} mainly draw inspirations from general object detection frameworks. TextBoxes~\cite{liao2017textboxes} adopted SSD~\cite{liu2016ssd} and added ``long'' default boxes and filters to handle the significant variation of aspect ratios of text instances. Based on Faster-RCNN~\cite{ren2015faster}, Ma~\textit{et al.}~\cite{ma2017arbitrary} devised Rotation Region Proposal Networks (RRPN) to detect arbitrary-Oriented text in natural images. EAST~\cite{Zhou_2017_CVPR} and Deep Regression~\cite{He_2017_ICCV} both directly produce the rotated boxes or quadrangles of text, in a per-pixel manner.

Segmentation based text detection methods cast text detection as a semantic segmentation problem and FCN~\cite{long2015fully} is often taken as the reference framework. Yao~\textit{et al.}~\cite{yao2016scene} modified FCN to produce multiple heatmaps corresponding various properties of text, such as text region and orientation. Zhang~\textit{et al.}~\cite{zhang2016multi} first use FCN to extract text blocks and then hunt character candidates from these blocks with MSER~\cite{neumann2010method}.  To better separate adjacent text instances, the method of~\cite{wu2017self} distinguishes each pixel into three categories: non-text, text border and text. These methods mainly vary in the way they separate text pixels into different instances.

The methods reviewed above have achieved excellent performances on various benchmarks in this field. However, most works, except for~\cite{yao2016scene,he2017multi,kheng2017total}, have not payed special attention to curved text. In contrast, the representation proposed in this paper is suitable for text of arbitrary shapes (horizontal, multi-oriented and curved). It is primarily inspired by~\cite{yao2016scene,he2017multi} and the geometric attributes of text are also estimated via the multiple-channel outputs  of an FCN-based model. Unlike~\cite{yao2016scene}, our algorithm does not need character level annotations. In addition, it also shares a similar idea with SegLink~\cite{Shi_2017_CVPR}, by successively decomposing text into local components and then composing them back into text instances. 
Analogous to~\cite{zhang2015symmetry}, we also detect linear symmetry axes of text instances for text localization.

Another advantage of the proposed method lies in its ability to reconstruct the precise shape and regional strike of text instances, which can largely facilitate the subsequent text recognition process, because all detected text instances could be conveniently transformed into a canonical form with minimal distortion and background (see the example in Fig.\ref{img_transform}).

\section{Methodology}
In this section, we first introduce the new representation for text of arbitrary shapes. Then we describe our method and training details.

\subsection{Representation} \label{sec:representation}

\begin{figure*}
\vspace{-6mm}
\begin{centering}
\includegraphics[width=0.75\columnwidth]{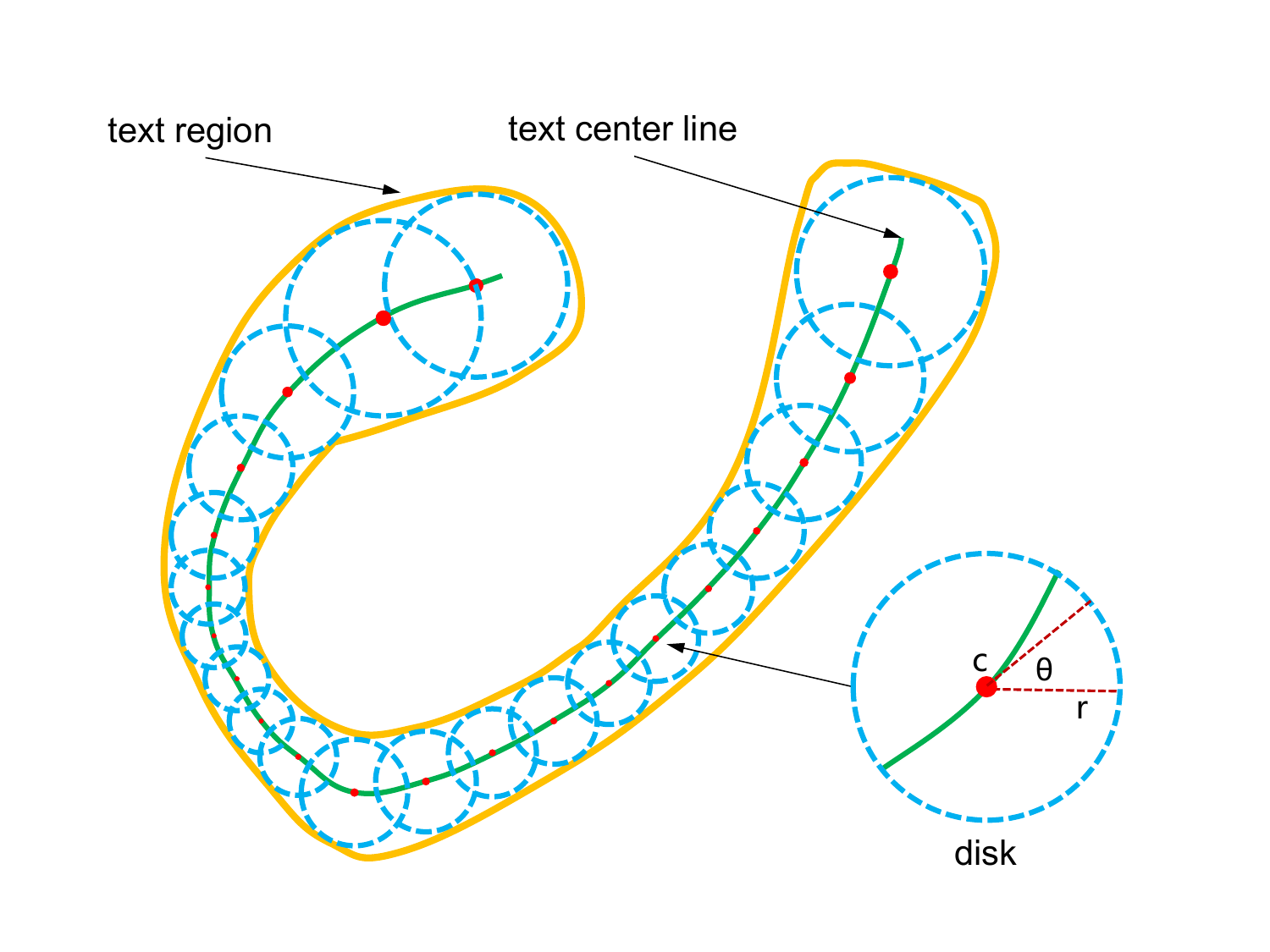}
\par\end{centering}
\vspace{-6mm}
\caption{Illustration of the proposed TextSnake representation. Text region (in yellow) is represented as a series of ordered disks (in blue), each of which is located at the center line (in green, a.k.a symmetric axis or skeleton) and associated with a radius $r$ and an orientation $\theta$. In contrast to conventional representations (e.g., axis-aligned rectangles, rotated rectangles and quadrangles), TextSnake is more flexible and general, since it can precisely describe text of different forms, regardless of shapes and lengths.}  \label{fig:textsnake}
\vspace{-5mm}
\end{figure*}

As shown in Fig.~\ref{fig:representations}, conventional representations for scene text (e.g., axis-aligned rectangles, rotated rectangles and quadrangles) fail to precisely describe the geometric properties of text instances of irregular shapes, since they generally assume that text instances are roughly in linear forms, which does not hold true for curved text. To address this problem, we propose a flexible and general representation: TextSnake. As demonstrated in Fig.~\ref{fig:textsnake}, TextSnake expresses a text instance as a sequence of overlapping disks, each of which is located at the center line and associated with a radius and an orientation. Intuitively, TextSnake is able to change its shape to adapt for the variations of text instances, such as rotation, scaling and bending.

Mathematically, a text instance $t$, consisting of several characters, can be viewed as an ordered list $S(t)$. $S(t)=\{D_{0},D_{1},\cdots,D_{i},\cdots,D_{n}\}$, where $D_{i}$ stands for the $i$th disk and $n$ is the number of the disks. Each disk $D$ is associated with a group of geometry attributes, i.e. $D=(c,r,\theta)$, in which $c$, $r$ and $\theta$ are the center, radius and orientation of disk $D$, respectively. The radius $r$ is defined as half of the local width of $t$, while the orientation $\theta$ is the tangential direction of the center line around the center $c$. In this sense, text region $t$ can be easily reconstructed by computing the union of the disks in $S(t)$.

Note that the disks do not correspond to the characters belonging to $t$. However, the geometric attributes in $S(t)$ can be used to rectify text instances of irregular shapes and transform them into rectangular, straight image regions, which are more friendly to text recognizers. 

\subsection{Pipeline} \label{sec:pipeline}

\begin{figure*}
\vspace{-3mm}
\begin{centering}
\includegraphics[width=0.9\columnwidth]{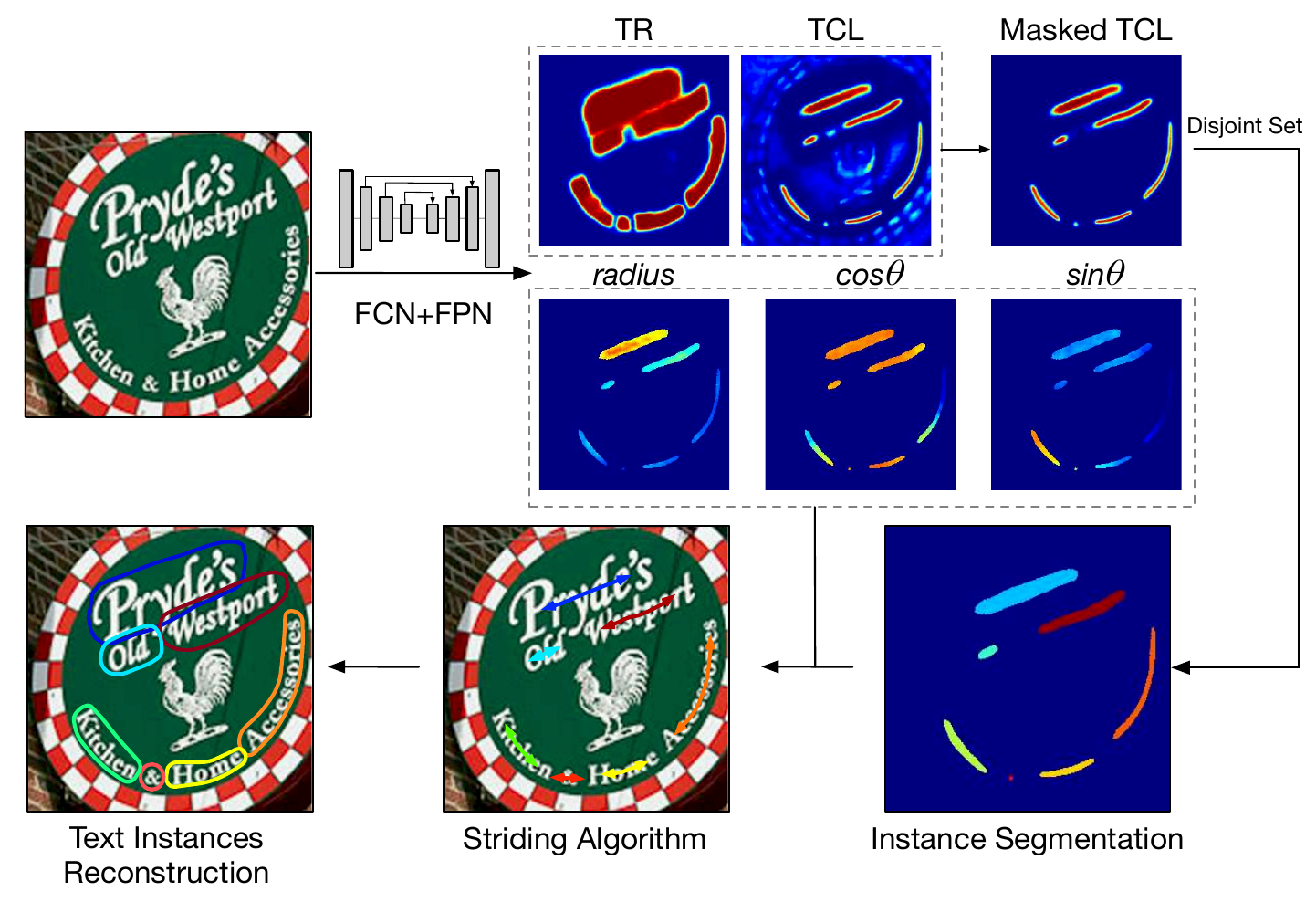}
\par\end{centering}
\vspace{-3mm}
\caption{Method framework: network output and post-processing}  \label{img_framework}
\vspace{-5mm}
\end{figure*}

In order to detect text with arbitrary shapes, we employ an FCN model to predict the geometry attributes of text instances. The pipeline of the proposed method is illustrated in Fig.\ref{img_framework}. The FCN based network predicts score maps of text center line (TCL) and  text regions (TR), together with geometry attributes, including $r$, $cos\theta$ and $sin\theta$. The TCL map is further masked by the TR map since TCL is naturally part of TR. To perform instance segmentation, disjoint set is utilized, given the fact that TCL does not overlap with each other. A striding algorithm is used to extract the central axis point lists and finally reconstruct the text instances.

\subsection{Network Architecture}

\begin{figure*}
\begin{centering}
\includegraphics[width=1.0\columnwidth]{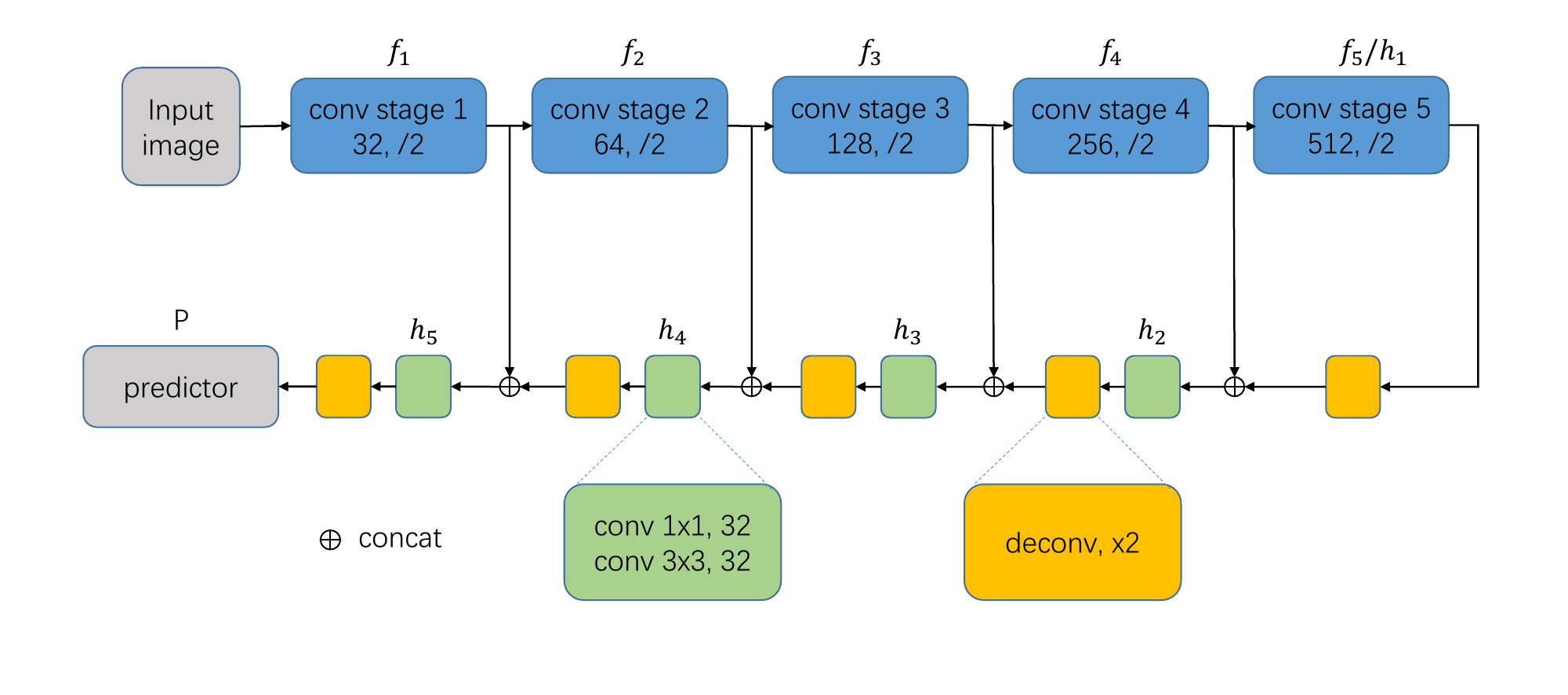}
\par\end{centering}
\vspace{-7mm}
\caption{Network Architecture. Blue blocks are convolution stages of VGG-16.}
\label{img_network}
\end{figure*}


The whole network is shown in Fig.~\ref{img_network}. Inspired by FPN\cite{Lin_2017_CVPR} and U-net\cite{Ronneberger2015U}, we adopt a scheme that gradually merges features from different levels of the stem network. The stem network can be convolutional networks proposed for image classification, e.g. VGG-16/19\cite{simonyan2014very} and ResNet\cite{He_2017_Res}. These networks can be divided into 5 stages of convolutions and a few additional fully-connected (FC) layers. We remove the FC layers, and feed the feature maps after each stage to the feature merging network. We choose VGG-16 as our stem network for the sake of direct and fair comparison with other methods. 

As for the feature merging network, several stages are stacked sequentially, each consisting of a merging unit that takes feature maps from the last stage and corresponding stem network layer. Merging unit is defined by the following equations:

\begin{equation}
h_1 = f_5
\label{eq1}
\end{equation}
\vspace{-2mm}
\begin{equation}
h_i = conv_{3\times 3}(conv_{1\times 1}{[f_{i-1}; UpSampling_{\times 2}(h_{i-1})]}),\  \mathrm{for} \ i\ge2
\label{eq2}
\end{equation}
\vspace{-2mm}

where $f_i$ denotes the feature maps of the $i$-th stage in the stem network and $h_i$ is the feature maps of the corresponding merging units. In our experiments, upsampling is implemented as deconvolutional layer as proposed in \cite{Zeiler2010Deconvolutional}. 


After the merging, we obtain a feature map whose size is $\frac{1}{2}$ of the input images. We apply an additional upsampling layer and 2 convolutional layers to produce dense predictions:

\begin{equation}
h_{final} = UpSampling_{\times 2}(h_5)
\label{eq3}
\end{equation}
\vspace{-2mm}
\begin{equation}
P = conv_{1\times 1}(conv_{3\times 3}(h_{final}))
\label{eq4}
\end{equation}
\vspace{-2mm}

where $P\in \mathcal{R}^{h\times w\times 7}$, with $4$ channels for logits of TR/TCL, and the last $3$ respectively for $r$, $cos\theta$ and $sin\theta$ of the text instance. As a result of the additional upsampling layer, $P$ has the same size as the input image.The final predictions are obtained by taking softmax for TR/TCL and regularizing $cos\theta$ and $sin\theta$ so that the squared sum equals $1$.

\subsection{Inference} \label{sec:inference}

After feed-forwarding, the network produces the TCL, TR and geometry maps. For TCL and TR, we apply thresholding with values $T_{tcl}$ and $T_{tr}$ respectively. Then, the intersection of TR and TCL gives the final prediction of TCL. Using disjoint-set, we can efficiently separate TCL pixels into different text instances.

Finally, a striding algorithm is designed to extract an ordered point list that indicates the shape and course of the text instance, and also reconstruct the text instance areas. Two simple heuristics are applied to filter out false positive text instances: 1) The number of TCL pixels should be at least $0.2$ times their average radius; 2) At least half of pixels in the reconstructed text area should be classified as TR.

\begin{figure*}
\vspace{-3mm}
\begin{centering}
\includegraphics[width=0.85\columnwidth]{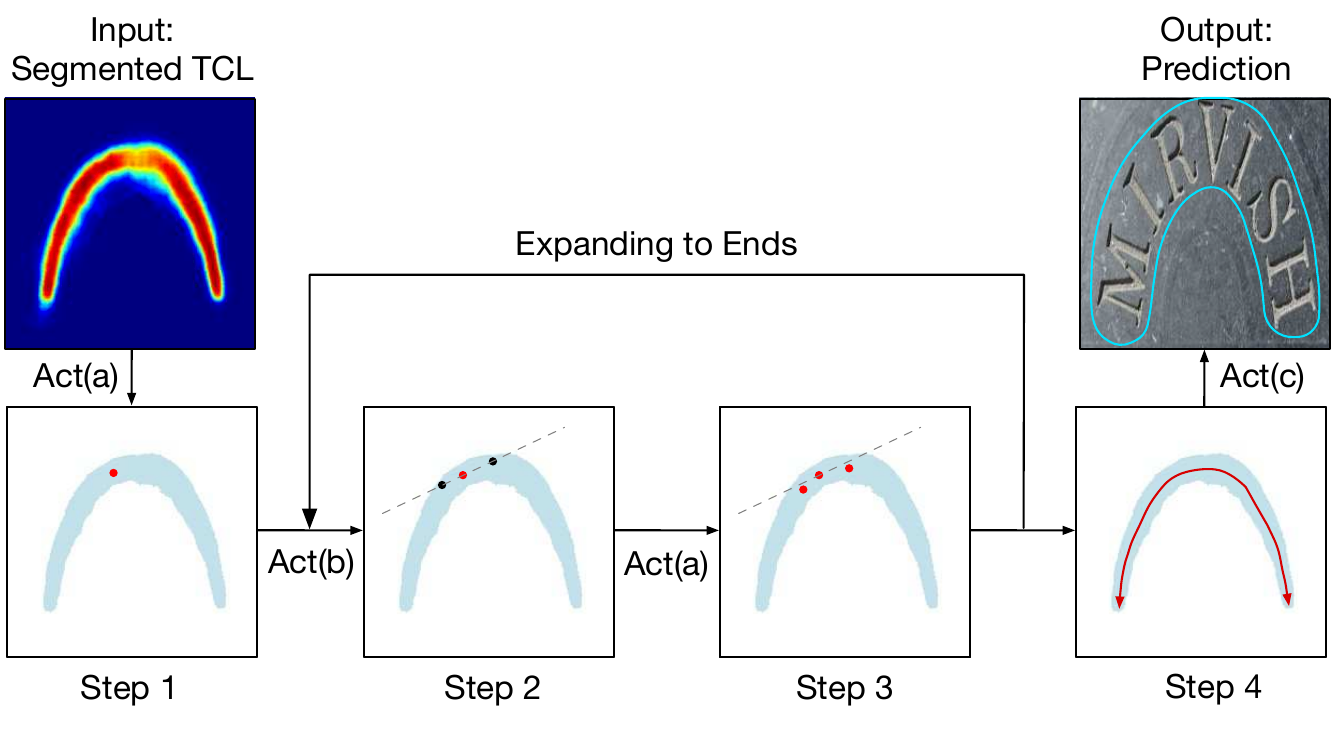}
\par\end{centering}
\vspace{-4mm}
\caption{Framework of Post-processing Algorithm. Act(a) Centralizing: relocate a given point to the central axis; Act(b) Striding: a directional search towards the ends of text instances; Act(c) Sliding: a reconstruction  by sliding a circle along the central axis.}
\label{img_PP}
\vspace{-3mm}
\end{figure*}

\begin{figure*}
\begin{centering}
\includegraphics[width=0.8\columnwidth]{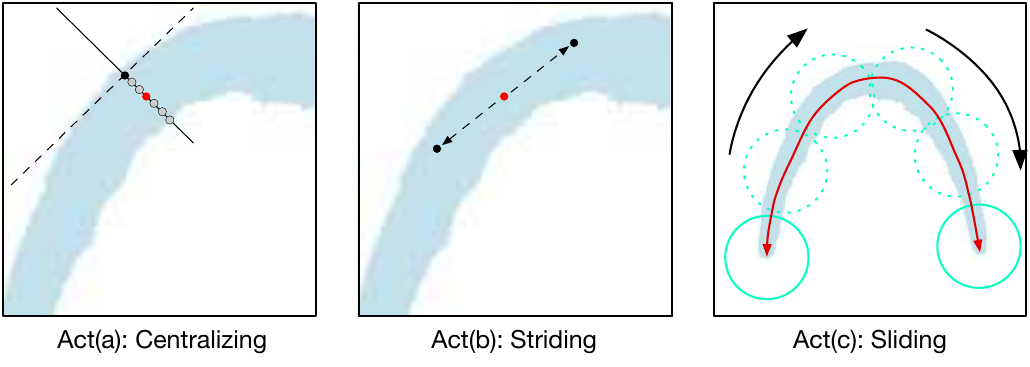}
\par\end{centering}
\vspace{-4mm}
\caption{Mechanisms of Centralizing, Striding and Sliding}
\label{img_PP_detail}
\end{figure*}

The procedure for the striding algorithm is shown in Fig.\ref{img_PP}. It features $3$ main  actions, denoted as Act(a), Act(b), and Act(c), as illustrated in Fig.\ref{img_PP_detail}. Firstly, we randomly select a pixel as the starting point, and centralize it. Then, the search process forks into two opposite directions, striding and centralizing until it reaches the ends. This process would generates 2 ordered point list in two opposite directions, which can be combined to produce the final central axis list that follows the course of the text and describe the shape precisely. Details of the $3$ actions are shown below.

\noindent\textbf{Act(a) Centralizing} As shown in Fig.\ref{img_PP_detail}, given a point on the TCL, we can draw the tangent line and the normal line, respectively denoted as dotted line and solid line.  This step can be done with ease using the geometry maps. The midpoint of the intersection of the normal line and the TCL area gives the centralized point.

\noindent\textbf{Act(b) Striding} The algorithm takes a stride to the next point to search. With the geometry maps, the displacement for each stride is computed and represented as $(\frac{1}{2} r \times cos\theta, \frac{1}{2}r \times sin\theta)$ and $(-\frac{1}{2}r \times cos\theta, -\frac{1}{2}r \times sin\theta)$, respectively for the two directions. If the next step is outside the TCL area, we decrement the stride gradually until it's inside, or it hits the ends.

\noindent\textbf{Act(c) Sliding} The algorithm iterates through the central axis and draw circles along it.  Radii of the circles are obtained from the $r$ map. The area covered by the circles indicates the predicted text instance.

In conclusion, taking advantage of the geometry maps and the TCL that precisely describes the course of the text instance, we can go beyond detection of text and also predict their shape and course. Besides, the striding algorithm saves our method from traversing all pixels that are related.

\subsection{Label Generation}
 
\subsubsection{Extracting Text Center Line}

For triangles and quadrangles, it's easy to directly calculate the TCL with algebraic methods, since in this case, TCL is a straight line. For polygons of more than 4 sides, it's not easy to derive a general algebraic method. 

Instead, we propose a method that is based on the assumption that, text instances are snake-shaped, i.e. that it does not fork into multiple branches. For a snake-shaped text instance, it has two edges that are respectively the \textit{head} and the \textit{tail}. The two edges near the head or tail are running parallel but in opposite direction.

\begin{figure*}
\begin{centering}
\includegraphics[width=1\columnwidth]{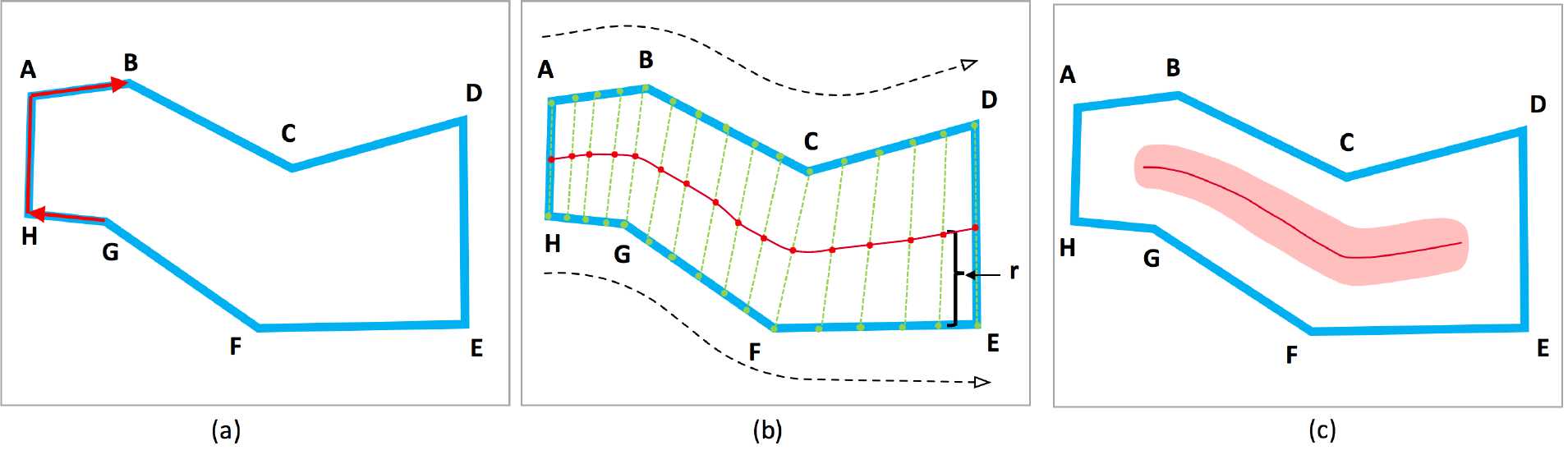}
\par\end{centering}
\vspace{-4mm}
\caption{Label Generation. (a) Determining text head and tail; (b) Extracting text center line and calculating geometries; (c) Expanded text center line.} \label{img_label}
\vspace{-5mm}
\end{figure*}

For a text instance $t$ represented by a group of  vertexes $\{v_0, v_1, v_2,...,v_n\}$ in clockwise or counterclockwise order, we define a  measurement for each edge $e_{i,i+1}$ as $M(e_{i,i+1})=\cos\langle e_{i+1,i+2}, e_{i-1,i}\rangle$. Intuitively, the two edges with $M$ nearest to $-1$, e.g. $AH$ and $DE$ in Fig.\ref{img_label}, are the head and tail. After that, equal number of anchor points are sampled on the two sidelines, e.g. $ABCD$ and $HGFE$ in Fig.\ref{img_label}. TCL points are computed as midpoints of corresponding anchor points. We shrink the two ends of TCL by $\frac{1}{2}r_{end}$ pixels, so that TCL are inside the TR and makes it easy for the network to learn to separate adjacent text instances. $r_{end}$ denotes the radius of the TCL points at the two ends. Finally, we expand the TCL area by $\frac{1}{5} r$, since a single-point line is prone to noise.





\subsubsection{Calculating $r$ and $\theta$}
For each points on TCL: (1) $r$ is computed as the distance to the corresponding point on sidelines; (2) $\theta$ is computed by fitting a straight line on the TCL points in the neighborhood. For non-TCL pixels, their corresponding geometry attributes are set to 0 for convenience. 

\subsection{Training Objectives}

The proposed model is trained end-to-end, with the following loss functions as the objectives:
\begin{equation}
L=L_{cls}+L_{reg}
\label{eq5}
\vspace{-4mm}
\end{equation}

\begin{equation}
L_{cls}=\lambda_{1}L_{tr} + \lambda_{2}L_{tcl}
\label{eq6}
\vspace{-4mm}
\end{equation}

\begin{equation}
L_{reg}=\lambda_{3}L_{r} +\lambda_{4}L_{sin} +\lambda_{5}L_{cos}
\label{eq7}
\end{equation}

$L_{cls}$ in Eq.\ref{eq5} represents classification loss for TR and TCL,  and $L_{reg}$ for regression loss of $r$, $cos\theta$ and $sin\theta$. In Eq.\ref{eq6}, $L_{tr}$ and $L_{tcl}$ are cross-entropy loss for TR and TCL. Online hard negative mining~\cite{Shrivastava2016Training} is adopted for TR loss, with the ratio between the negatives and positives kept to 3:1 at most. For TCL, we only take into account pixels inside TR and adopt no balancing methods.

In Eq.\ref{eq7}, regression loss, i.e. $L_{r}$ $L_{sin}$ and $L_{cos}$, are calculated as Smoothed-L1 loss\cite{Girshick_2015_ICCV}:

\begin{equation}
\begin{pmatrix}L_{r} \\ L_{cos}  \\ L_{sin}\end{pmatrix}=
SmoothedL1\begin{pmatrix}\frac{\widehat{r}-r}{r} \\ 
\widehat{cos\theta}-cos\theta  \\
\widehat{sin\theta}-sin\theta\end{pmatrix}
\label{eq8}
\end{equation}

where $\widehat r$, $\widehat{cos\theta}$ and $\widehat{sin\theta}$ are the predicted values, while $r$, $cos\theta$ and $sin\theta$ are their ground truth correspondingly. Geometry loss outside TCL are set to 0, since these attributes make no sense for non-TCL points.

The weights constants $\lambda_{1}$, $\lambda_{2}$, $\lambda_{3}$, $\lambda_{4}$ and $\lambda_{5}$ are all set to  1 in our experiments.


\section{Experiments}

In this section, we evaluate the proposed algorithm on standard benchmarks for scene text detection and compare it with previous methods. Analyses and discussions regarding our algorithm are also given.

\subsection{Datasets}

The datasets used for the experiments in this paper are briefly introduced below:

\textbf{SynthText}~\cite{gupta2016synthetic} is a large sacle dataset that contains about $800K$ synthetic images. These images are created by blending natural images with text rendered with random fonts, sizes, colors, and orientations, thus these images are quite realistic. We use this dataset to pre-train our model.


\textbf{TotalText}~\cite{kheng2017total} is a newly-released benchmark for text detection. Besides horizontal and multi-Oriented text instances, the dataset specially features~\textit{curved text}, which rarely appear in other benchmark datasets,but are actually quite common in real environments. The dataset is split into training and testing sets with 1255 and 300 images, respectively.

\textbf{CTW1500}~\cite{Yuliang2017Detecting} is another dataset mainly consisting of curved text. It consists of 1000 training images and 500 test images. Text instances are annotated with polygons with 14 vertexes.

\textbf{ICDAR 2015} is proposed as the Challenge 4 of the 2015 Robust Reading Competition \cite{karatzas2015icdar} for incidental scene text detection. Scene text images in this dataset are taken by Google Glasses without taking care of positioning, image quality, and viewpoint. This dataset features small, blur, and multi-oriented text instances. There are 1000 images for training  and 500 images for testing. The text instances from this dataset are labeled as word level quadrangles. 

\textbf{MSRA-TD500}~\cite{yao2012detecting} is a dataset with multi-lingual, arbitrary-oriented and long text lines. It includes 300 training images and 200 test images with text line level annotations. Following previous works~\cite{Zhou_2017_CVPR,Lyu2018}, we also include the images from HUST-TR400~\cite{yao2014unified} as training data when fine-tuning on this dataset, since its training set is rather small. 

For experiments on ICDAR 2015 and MSRA-TD500, we fit a minimum bounding rectangle based on the output text area of our method.

\subsection{Data Augmentation}

Images are randomly rotated, and cropped with areas ranging from $0.24$ to $1.69$ and aspect ratios ranging from $0.33$ to $3$. After that, noise, blur, and lightness are randomly adjusted. We ensure that the text on the augmented images are still legible, if they are legible before augmentation.

\begin{figure*}
\begin{centering}
\includegraphics[width=0.95\columnwidth]{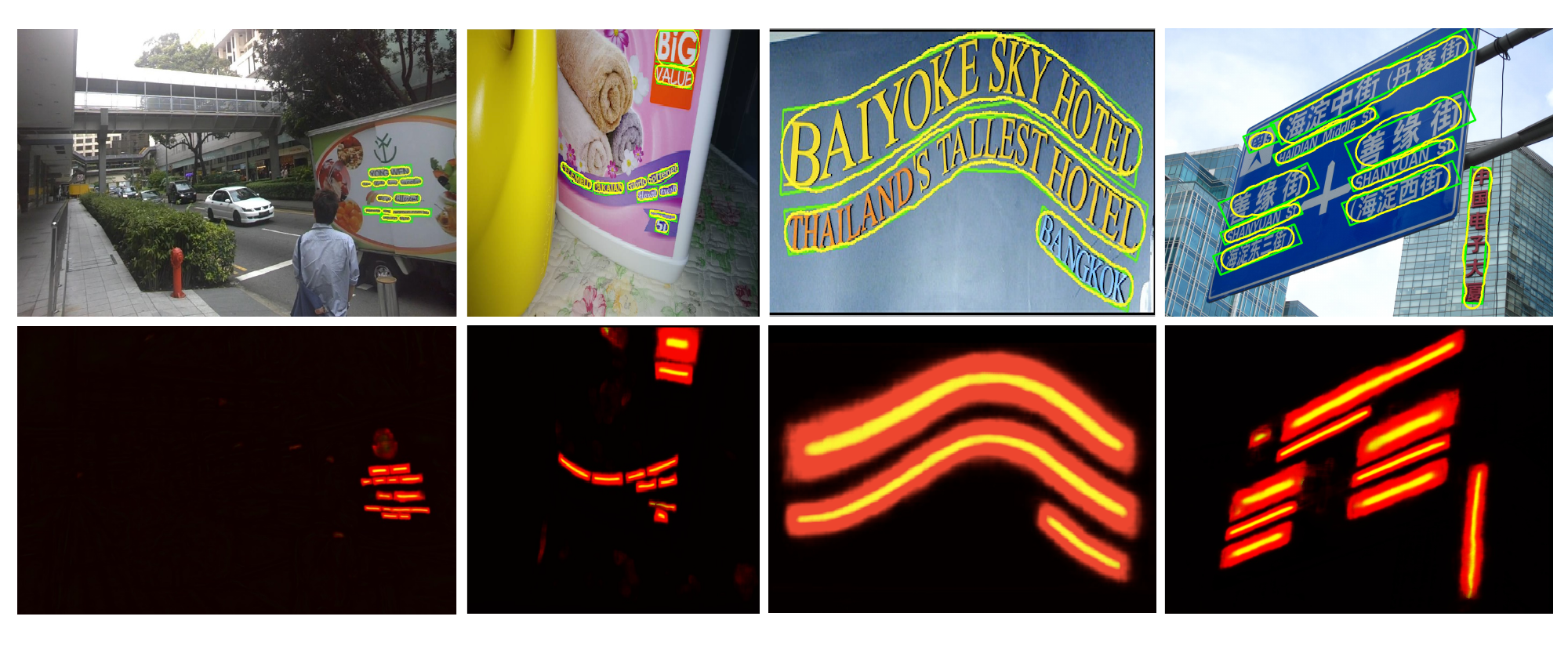}
\par\end{centering}
\vspace{-5mm}
\caption{Qualitative results by the proposed method. Top: Detected text contours (in \textit{yellow}) and ground truth annotations (in \textit{green}). Bottom: Combined score maps for TR (in \textit{red}) and TCL (in \textit{yellow}). From left to right in column: image from ICDAR 2015, TotalText, CTW1500 and MSRA-TD500. Best viewed in color.}
\label{img_samples}
\vspace{-3mm}
\end{figure*}

\subsection{Implementation Details}

Our method is implemented in Tensorflow 1.3.0~\cite{abadi2016tensorflow}.
The network is pre-trained on SynthText for one epoch and fine-tuned on other datasets. We adopt the Adam optimazer~\cite{kingma2014adam} as our learning rate scheme. During the pre-training stage, the learning rate is fixed to $10^{-3}$. During the fine-tuning stage, the learing rate is set to $10^{-3}$ initially and decaies with a rate of $0.8$ every 5000 iterations. During fine-tuning, the number of iterations is decided by the sizes of datasets. All the experiments are conducted on a regular workstation (CPU: Intel(R) Xeon(R) CPU E5-2650 v3 @ 2.30GHz; GPU:Titan X; RAM: 384GB). We train our model with the batch size of 32 on $2$ GPUs in parallel and evaluate our model on 1 GPU with batch size set as $1$. Hyper-parameters are tuned by grid search on training set.

\subsection{Experiment Results}

\subsubsection*{Experiments on Curved Text (Total-Text and CTW1500)}

Fine-tuning on these two datasets stops at about $5k$ iterations. Thresholds $T_{tr}$, $T_{tcl}$ are set to $(0.4, 0.6)$ and $(0.4, 0.5)$ respectively on Total-Text and CTW1500. 
In testing, all images are rescaled to $512\times 512$ for Total-Text, while for CTW1500, the images are not resized, since the images in CTW1500 are rather small (The largest image is merely $400\times 600$). For comparison, we also evaluated the models of EAST~\cite{Zhou_2017_CVPR} and SegLink~\cite{Shi_2017_CVPR} on Total-Text and CTW1500. The quantitative results of different methods on these two datasets are shown in Tab.~\ref{tab_total} and Tab.~\ref{tab_CTW1500}, respectively. 
\begin{table}
\small
\begin{centering}
\vspace{-2mm}
\caption{Quantitative results of different methods evaluated on Total-Text. Note that EAST and SegLink were not fine-tuned on Total-Text. Therefore their results are included only for reference.
}
\label{tab_total}
\begin{tabular}{|c|c|c|c|}
\hline 
\textbf{Method} & \textbf{Precision} & \textbf{Recall} & \textbf{F-measure} \tabularnewline
\hline 
\hline 
SegLink \cite{Shi_2017_CVPR} & $30.3$  & $23.8$ & $26.7$\tabularnewline
\hline 
EAST \cite{Zhou_2017_CVPR} & $50.0$  & $36.2$ & $42.0$ \tabularnewline
\hline 
Baseline (DeconvNet\cite{Noh2015Learning}) & $33.0$ & $40.0$ & $36.0$ \tabularnewline
\hline
\textbf{TextSnake} & \textbf{82.7} & \textbf{74.5} & \textbf{78.4} \tabularnewline
\hline
\end{tabular}
\par\end{centering}
\end{table}

As shown in Tab. \ref{tab_total}
, the proposed method achieves $82.7\%$, $74.5\%$, and $78.4\%$ in precision, recall and F-measure on Total-Text, significantly outperforming previous methods. Note that the F-measure of our method is more than double of that of the baseline provided in the original Total-Text paper~\cite{kheng2017total}.
\begin{table}
\small
\vspace{-2mm}
\begin{centering}
\caption{Quantitative results of different methods evaluated on CTW1500. Results other than ours are obtained from \cite{Yuliang2017Detecting}.} \label{tab_CTW1500}
\begin{tabular}{|c|c|c|c|}
\hline 
\textbf{Method} & \textbf{Precision} & \textbf{Recall} & \textbf{F-measure} \tabularnewline
\hline 
\hline 
SegLink \cite{Shi_2017_CVPR} & $42.3$  & $40.0$ & $40.8$ \tabularnewline
\hline 
EAST \cite{Zhou_2017_CVPR} & $78.7$  & $49.1$ & $60.4$ \tabularnewline
\hline 
DMPNet \cite{Liu2017Deep} & $69.9$  & $56.0$ & $62.2$ \tabularnewline
\hline 
CTD\cite{Yuliang2017Detecting} & $74.3$ & $65.2$ & $69.5$ \tabularnewline
\hline 
CTD+TLOC\cite{Yuliang2017Detecting} & \textbf{77.4} & $69.8$ & $73.4$ \tabularnewline
\hline
\textbf{TextSnake} & 67.9 & \textbf{85.3} & \textbf{75.6} \tabularnewline
\hline 
\end{tabular}
\par\end{centering}
\vspace{-4mm}
\end{table}

On CTW1500, the proposed method achieves $67.9\%$, $85.3\%$, and $75.6\%$ in precision, recall and F-measure
, respectively. Compared with CTD+TLOC which is proposed together with the CTW1500 dataset in
~\cite{Yuliang2017Detecting}, the F-measure of our algorithm is $2.2\%$ higher ($75.6\%$ vs. $73.4\%$).

The superior performances of our method on Total-Text and CTW1500 verify that the proposed representation can handle well curved text in natural images.

\subsubsection*{Experiments on Incidental Scene Text (ICDAR 2015)}

Fine-tuning on ICDAR 2015 stops at about $30k$ iterations.
In testing, all images are resized to $1280\times 768$. $T_{tr}$, $T_{tcl}$ are set to $(0.4, 0.9)$.  For the consideration that images in ICDAR 2015 contains many unlabeled small texts,  predicted rectangles with the shorter side less than 10 pixels or the area less than 300 are filtered out. 

The quantitative results of different methods on ICDAR 2015 are shown in Tab.\ref{tab_icdar2015}. 
With only \textit{single-scale} testing, our method outperforms most competitors (including those evaluated in multi-scale). 
This demonstrates that the proposed representation TextSnake is general and can be readily applied to multi-oriented text in complex scenarios.
\begin{table}
\small
\begin{centering}
\vspace{-2mm}
\caption{Quantitative results of different methods on ICDAR 2015. $^*$ stands for multi-scale, $^\text{\dag}$ indicates that the base net of the model is not VGG16.}  \label{tab_icdar2015}
\begin{tabular}{|c|c|c|c|c|}
\hline 
\textbf{Method} & \textbf{Precision} & \textbf{Recall} & \textbf{F-measure} & \textbf{FPS} \tabularnewline
\hline 
\hline 
Zhang \emph{et al.} \cite{zhang2016multi} & 70.8 & 43.0 & 53.6 & 0.48 \tabularnewline
\hline 
CTPN \cite{tian2016detecting} & 74.2  & 51.6  & 60.9 & 7.1 \tabularnewline
\hline 
Yao \emph{et al.} \cite{yao2016scene} & 72.3  & 58.7  & 64.8 & 1.61 \tabularnewline
\hline 
SegLink \cite{Shi_2017_CVPR} & 73.1  & 76.8 & 75.0 & - \tabularnewline
\hline 
EAST \cite{Zhou_2017_CVPR} & 80.5  & 72.8 & 76.4 & 6.52\tabularnewline
\hline 
SSTD \cite{SSTD} & 80.0  & 73.0 & 77.0 & \textbf{7.7} \tabularnewline
\hline
WordSup $^*$ \cite{Hu_2017_ICCV} & 79.3  & 77.0  & 78.2 & 2 \tabularnewline
\hline 
EAST $^*$ $^\text{\dag}$ \cite{Zhou_2017_CVPR}& 83.3  & 78.3 & 80.7 & - \tabularnewline
\hline 
He \emph{et al.} $^*$ $^\text{\dag}$ \cite{He_2017_ICCV} & 82.0  &  80.0 & 81.0 & 1.1 \tabularnewline
\hline
PixelLink \cite{deng2018pixellink} & \textbf{85.5}  & \textbf{82.0} & \textbf{83.7} & 3.0 \tabularnewline
\hline
\textbf{TextSnake} & 84.9 & 80.4 & 82.6 & 1.1\tabularnewline
\hline
\end{tabular}
\par\end{centering}
\vspace{-6mm}
\end{table}

\subsubsection*{Experiments on Long Straight Text Lines (MSRA-TD500)}

Fine-tuning on MSRA-TD500 stops at about $10k$ iterations. Thresholds for $T_{tr}$, $T_{tcl}$ are $(0.4, 0.6)$ . 
In testing, all images are resized to $1280\times 768$. Results are shown in Tab.\ref{tab_td500}. The F-measure ($78.3\%$) of the proposed method is higher than that of the other methods.

\begin{table}
\small
\begin{centering}
\caption{Quantitative results of different methods on MSRA-TD500. $^\text{\dag}$ indicates models whose base nets are not VGG16.}  \label{tab_td500}
\begin{tabular}{|c|c|c|c|c|}
\hline 
\textbf{Method} & \textbf{Precision} & \textbf{Recall} & \textbf{F-measure} &\textbf{FPS} \tabularnewline
\hline 
\hline
Kang \emph{et al.} \cite{kang2014orientation} & 71.0 &  62.0 & 66.0 & -\tabularnewline
\hline
Zhang \emph{et al.} \cite{zhang2016multi} & 83.0 & 67.0 & 74.0 & 0.48\tabularnewline
\hline 
Yao \emph{et al.} \cite{yao2016scene} & 76.5  & \textbf{75.3}  & 75.9 &  1.61\tabularnewline
\hline 
EAST \cite{Zhou_2017_CVPR}  & 81.7  & 61.6 & 70.2 & 6.52 \tabularnewline
\hline 
EAST  $^\text{\dag}$ \cite{Zhou_2017_CVPR} & \textbf{87.3}  & 67.4 & 76.1 & \textbf{13.2} \tabularnewline
\hline 
SegLink \cite{Shi_2017_CVPR} & 86.0  & 70.0 & 77.0 & 8.9 \tabularnewline
\hline 
He \emph{et al.} $^\text{\dag}$ \cite{He_2017_ICCV} & 77.0  &  70.0 & 74.0 & 1.1 \tabularnewline
\hline
PixelLink \cite{deng2018pixellink} & 83.0  & 73.2 & 77.8 & 3.0 \tabularnewline
\hline
\textbf{TextSnake} & 83.2 & 73.9 & \textbf{78.3} & 1.1 \tabularnewline
\hline
\end{tabular}
\par\end{centering}
\vspace{-3mm}
\end{table}

\subsection{Analyses and Discussions}


\noindent\textbf{Precise Description of Text Instances} What distinguishes our method from others is its ability to predict a precise description of the shape and course of text instances(see Fig.\ref{img_samples}).

We attribute such ability to the TCL mechanism. Text center line can be seen as a kind of skeletons that prop up the text instance, and geo-attributes providing more details. Text, as a form of written language, can be seen as a stream of signals mapped onto 2D surfaces. Naturally, it should follows a course to extend.

Therefore we propose to predict TCL, which is much narrower than the whole text instance. It has two advantages: (1) A slim TCL can better describe the course and shape; (2) TCL, intuitively, does not overlaps with each other, so that instance segmentation can be done in a very simple and straightforward way, thus simplifying our pipeline.

\begin{figure*}
\begin{centering}
\vspace{-3mm}
\includegraphics[width=1.0\columnwidth]{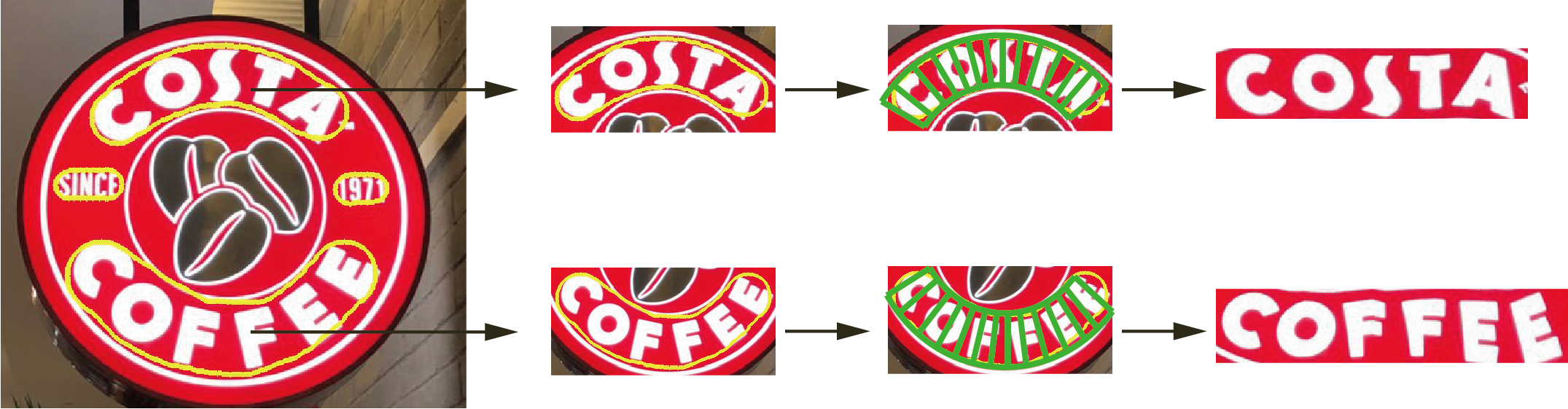}
\par\end{centering}
\vspace{-3mm}
\caption{Text instances transformed to canonical form using the predicted geometries.} \label{img_transform}
\vspace{-0mm}
\end{figure*}

Moreover, as depicted in Fig.\ref{img_transform}, we can exploit local geometries to sketch the structure of the text instance and transform the predicted curved text instances into canonical form, which may largely facilitate the recognition stage.

\noindent\textbf{Generalization Ability} To further verify the generalization ability of our method, we train and fine-tune our model on datasets \textit{without} curved text and evaluate it on the two benchmarks \textit{featuring} curved text. Specifically, we fine-tune our models on ICDAR 2015, and evaluate them on the target datasets. The models of EAST~\cite{Zhou_2017_CVPR},  SegLink~\cite{Shi_2017_CVPR}, and PixelLink~\cite{deng2018pixellink} are taken as baselines, since these two methods were also trained on ICDAR 2015. 

\begin{table}
\small
\begin{centering}
\caption{Comparison of cross-dataset results of different methods. The following models are fine-tuned on ICDAR 2015 and evaluated on Total-Text and CTW1500. Experiments for SegLink, EAST and PixelLink are done with the open source code. The evaluation protocol is DetEval~\cite{wolf2006object}, the same as Total-Text. While ICDAR 2015 and Total-Text has word-level labels, CTW1500 uses line-level ones. We deem DetEval\cite{wolf2006object} preferable to PASCAL \cite{everingham2015pascal}. Otherwise, the line-level labels of CTW1500 would significantly penalize models fine-tuned on word-level labeled ICDAR2015.} 

\label{tab_cross}
\begin{tabular}{|c|c|c|c|c|c|c|}
\hline 
{\textbf{Datasets}} & \multicolumn{3}{|c|}{\textbf{Total-Text}} & \multicolumn{3}{|c|}{\textbf{CTW1500}} \tabularnewline
\hline
\textbf{Methods} & \textbf{Precision} & \textbf{Recall} & \textbf{F-measure} & \textbf{Precision} & \textbf{Recall} & \textbf{F-measure}\tabularnewline

\hline
SegLink\cite{Shi_2017_CVPR} & 35.6 & 33.2 & 34.4 & 33.0& 28.4& 30.5\tabularnewline
\hline
EAST\cite{Zhou_2017_CVPR} & 49.0 & 43.1 & 45.9 & 46.7& 37.2& 41.4\tabularnewline
\hline
PixelLink~\cite{deng2018pixellink} & 53.5 & 52.7 & 53.1 & 50.6& 42.8& 46.4\tabularnewline
\hline
\hline
\textbf{TextSnake} & $\textbf{61.5}$ & \textbf{67.9} & \textbf{64.6} & \textbf{65.4}& \textbf{63.4}& \textbf{64.4} \tabularnewline
\hline
\end{tabular}
\par\end{centering}
\vspace{-2mm}
\end{table}

As shown in Tab.\ref{tab_cross}, our method still performs well on curved text and significantly outperforms the three strong competitors SegLink, EAST and PixelLink, without fine-tuning on curved text. We attribute this excellent generalization ability to the proposed flexible representation. Instead of taking text as a whole, the representation treats text as a collection of local elements and integrates them together to make decisions. Local attributes are kept when formed into a whole. Besides, they are independent of each other. Therefore, the final predictions of our method can retain most information of the shape and course of the text.We believe that this is the main reason for the capacity of the proposed text detection algorithm in hunting text instances with various shapes.





\section{Conclusion and Future Work}

In this paper, we present a novel, flexible representation for describing the properties of scene text with arbitrary shapes, including horizontal, multi-oriented and curved text instances. The proposed text detection method based upon this representation obtains state-of-the-art or comparable performance on two newly-released benchmarks for curved text (Total-Text and SCUT-CTW1500) as well as two widely-used datasets (ICDAR 2015 and MSRA-TD500) in this field, proving the effectiveness of the proposed method. As for future work, we would explore the direction of developing an end-to-end recognition system for text of arbitrary shapes.




\bibliographystyle{splncs}
\bibliography{TextSnake.bib}
\end{document}